\documentclass[letterpaper, 10 pt, conference]{ieeeconf}  % Comment this line out if you need a4paper

\IEEEoverridecommandlockouts                              % This command is only needed if 
\usepackage{amsmath} % assumes amsmath package installed
\usepackage{amssymb}  % assumes amsmath package installed

\usepackage{cite}
\usepackage{algorithmic}
\usepackage{graphicx}
\usepackage{textcomp}
\usepackage[svgnames]{xcolor}
\usepackage{bm}
\usepackage{multirow}
\usepackage{siunitx}
\usepackage{tikz, pgfplots}
\usepackage{mathtools}
\usepackage{verbatim}
\usepackage{babel}
\usepackage{filecontents}
\usepackage{pgfplotstable}
\usepgfplotslibrary{colorbrewer}
\pgfplotsset{compat = 1.15, cycle list/Set1-8} 
\usetikzlibrary{pgfplots.statistics, pgfplots.colorbrewer} 
\usepackage{filecontents}
\usepackage{float} 
\usepackage{graphicx}
\usepackage{caption}
\usepackage{subcaption}

\usepackage{ifthen}
\usepackage{pgfplots}
\usepackage{subcaption}
\graphicspath{{./fig/}}
\usetikzlibrary{arrows, calc,intersections, shapes}
\usetikzlibrary{arrows.meta}
\usetikzlibrary{decorations.pathmorphing}
\usetikzlibrary{decorations.text}
\usetikzlibrary{decorations.pathreplacing}
\usetikzlibrary{hobby,decorations.markings}
\usetikzlibrary{positioning}
\usetikzlibrary{calc}
\tikzset{%
	every neuron/.style={
		circle,
		draw,
		minimum size=0.15cm
	},
	neuron missing/.style={
		draw=none, 
		scale=1,
		text height=1,
		execute at begin node=\color{black}$\vdots$
	},
}

\usepackage[switch]{lineno}  
\usepackage[hyphens]{url}  % DO NOT CHANGE THIS
\def\BibTeX{{\rm B\kern-.05em{\sc i\kern-.025em b}\kern-.08em
		T\kern-.1667em\lower.7ex\hbox{E}\kern-.125emX}}

\usepackage{textcomp}
\newcommand\copyrighttext{%
	\footnotesize \textcopyright 2023 IEEE. Personal use of this material is permitted. Permission from IEEE must be obtained for all other uses, in any current or future media, including reprinting/republishing this material for advertising or promotional purposes, creating new collective works, for resale or redistribution to servers or lists, or reuse of any copyrighted component of this work in other works. DOI: t.\,b.\,d.}
\newcommand\copyrightnotice{%
	\begin{tikzpicture}[remember picture,overlay]
	\node[anchor=south,yshift=10pt] at (current page.south) {\fbox{\parbox{\dimexpr\textwidth-\fboxsep-\fboxrule\relax}{\copyrighttext}}};
	\end{tikzpicture}%
}
\begin{document}
\title{\LARGE \bf
Optimization and Interpretability of Graph Attention Networks for Small Sparse Graph Structures in Automotive Applications}

\author{
	Marion Neumeier\textsuperscript{\rm 1}*\thanks{*We appreciate the funding of this work by AUDI AG.},
	Andreas Tollkühn\textsuperscript{\rm 2},
	Sebastian Dorn\textsuperscript{\rm 2},
	Michael Botsch\textsuperscript{\rm 1} and 
	Wolfgang Utschick\textsuperscript{\rm 3}\\	
	\thanks{\textsuperscript{\rm 1} CARISSMA Institute of Automated Driving, Technische Hochschule Ingolstadt, 85049 Ingolstadt, Germany {\tt\small firstname.lastname@thi.de}}%
	\thanks{\textsuperscript{\rm 2} AUDI AG, 85057 Ingolstadt, Germany {\tt\small firstname.lastname@audi.de}}
	\thanks{\textsuperscript{\rm 3} Technical University of Munich, 80333 Munich, Germany {\tt\small utschick@tum.de}}
}
\maketitle
\copyrightnotice
%%%%%%%%%%%%%%%%%%%%%%%%%%%%%%%%%%%%%%%%%%%%%%%%%%%%%%%%%%%%%%%%%%%%%%%%%%%%%%%
\def \vneg {-0.35cm}
%%%%%%%% ABSTRACT
\begin{abstract}
	 For automotive applications, the Graph Attention Network (GAT) is a prominently used architecture to include relational information of a traffic scenario during feature embedding. As shown in this work, however, one of the most popular GAT realizations, namely GATv2, has potential pitfalls that hinder an optimal parameter learning. Especially for small and sparse graph structures a proper optimization is problematic. To surpass limitations, this work proposes architectural modifications of GATv2. In controlled experiments, it is shown that the proposed model adaptions improve prediction performance in a node-level regression task and make it more robust to parameter initialization. This work aims for a better understanding \mbox{of the} attention mechanism and analyzes its interpretability of identifying causal importance. 
\end{abstract}
\section{Introduction}
In recent years, Graph Neural Networks (GNNs) have gained increasing popularity. GNNs are a class of deep learning architectures that leverage the representation of graph-structured data by incorporating the structure into the embedding. An eminent GNN realization is the Graph Attention Network (GAT) \cite{Velickovic.30.10.2017}, which performs attentional message-passing. Similar to the attention mechanism for sequential data, GATs update node features of a graph by weighting the features of its neighboring nodes using assigned attention scores. If designed and trained properly, the attention mechanism can improve a network’s robustness by attending to only important nodes in graph-structured data\cite{understandGAT}. Due to the inherent selection characteristic of the attention mechanism, several works claim that high attention scores indicate highly relevant information and conclude a certain interpretability \cite{Velickovic.30.10.2017}\cite{Park.2017}\cite{Deac.2019}\cite{Renz.25.10.2022}. However, recent works on sequential data using the attention mechanism have questioned this type of interpretability\cite{Hassid.07.11.2022}\cite{Serrano.2019}. Results suggest that the input-depended computation of attention matrices is less important than typically thought. While attention mechanisms have shown to be very powerful architectural components, there is dissent about how much they actually learn to attend and select relevant information. In general, the interpretability of data-driven architectures is a significant research topic\cite{Neumeier.2021}.

While most state-of-the-art prediction methods in the automotive field are based on conventional machine learning approaches, graph-based approaches have become very popular \cite{Neumeier.GFTNN}\cite{Klimke.642022692022} \cite{Liu.2022} \cite{Naik.642022692022}. Representing a traffic scenario as graph-structured data allows to include relational information, which GNN architectures are able to inherently incorporate into the embedding. Particularly, the ability of GATs to weigh different relations context-adaptively seems to be very beneficial and could potentially identify relevant inter-dependencies\cite{GATv2}. Beside the very prominent example of graph-based attentional aggregation in VectorNet \cite{VectorNET.2020}, various works use GATs to model high-level interactions of agent trajectories and/or map features \cite{Mo.2022} \cite{Liu.2022}\cite{xu2018how}. In comparison to other application fields like, e.\,g., brain analysis, the graph structures in automotive applications are rather small and sparse. As shown in this work, however, especially small and sparse graph architectures are likely to hinder appropriate parameter optimization in the GAT such that even simple graph problems cannot be solved reliably. To be aware of these constrains in GATs and know how to mitigate these effects is therefore of great importance.
% It reveals the reason, why the model is especially sensitive to parameter initialization.
With focus on the application in the automotive domain, this work identifies key weaknesses of one of the most relevant GAT realizations, namely GATv2\cite{GATv2}, and proposes architectural changes to surpass limitations. Experiments show, that the model adaptions improve performance and reinforce robustness to parameter initialization. Additionally, the attention assignment of the GAT is investigated with regard to (w.\,r\,.t.) its interpretability as relevance indication. In order to evaluate if the attention mechanism learns to generalize relevant dependencies within a graph structure, ground truth data of the attention scores are required. In real-world datasets, however, this information is hardly accessible. Such annotations are extremely hard to define and, if possible, require expensive labeling procedures. For this work, therefore, a synthetic graph dataset is generated. The simulated dataset allows to evaluate the attention mechanism in a controlled environment.
%A highly relevant information for a traffic participant, for example, could be a traffic sign.

\textbf{Contribution.} This work contributes towards a better understanding of graph-based attentional information aggregation.
The main contributions are as follows:
\begin{itemize}
	\item Identification of potential pitfalls of GATv2\cite{GATv2} with focus on automotive applications.
	\item Proposition of architectural modifications to surpass limitations.
	\item Evaluation of performance and interpretability of the graph-based attention mechanism based on controlled experiments. 
\end{itemize}
In this work, vectors are denoted as bold lowercase letters and matrices as bold capital letters.
\section{Related Work}
This section reviews related work w.\,r.\,t. the expressive power of GNNs and, in particular, GATs, as well as the understanding and interpretability of attention mechanisms.
\subsection{Expressive power of GNNs}
Over the past years, several works have addressed the discriminative power of many common and popular GNN frameworks \cite{xu2018how} \cite{BruelGabrielsson.2020} \cite{understandGAT} \cite{Mo.2022}. It is well known that GNNs generally are prone to oversmoothing, which gradually decreases their performance with increasing number of layers\cite{Chen.2020}. For this reason, most GNN implementations are kept shallow in practice. Subsequently, standalone layers should be maximally powerful and provide beneficial filter characteristics. However, Balcilar \textit{et al.}~\cite{Balcilar.2021} showed that most GNNs operate as a low-pass filter purely and only few networks allow bandpass-filtering. In order to leverage the bandpass-filtering capacity, numerous computational layers are required. Yet, this results in a trade-off between avoiding oversmooting issues and computational power. GATs have shown to be able to alleviate oversmoothing through the attentive message passing operations\cite{Min.2020}. If the attention mechanism generalizes to operate selectively instead of solely averaging neighboring nodes, oversmoothing can be prevented. While there are several approaches to compute attention scores in graphs, the mechanism GATv2 presented by Brody \textit{et al.} \cite{GATv2} is strictly more expressive than others \cite{Velickovic.30.10.2017}\cite{Vaswani.2017}. Furthermore, it has been shown that the popular GAT architecture proposed in \cite{Velickovic.30.10.2017} only performs a limited kind of attention, where the ranking of the attention scores is unconditional on the query node. 

\subsection{Interpretability of the attention mechanism}
Similar to the attention mechanism for sequential data \cite{Luong.17.08.2015}\cite{Vaswani.2017}, graph attention is supposed to capture important relations in graphs. Meaning, the attention score assigned to a neighboring node in a graph indicates the degree of its importance w.\,r.\,t. the reference node. In the most common approaches, every node attends to its neighbors by considering its own representation as a query. Although various works have leveraged performance by using the GAT, the degree of performance improvements are inconsistent across datasets and it's still not fully understood what the graph attention mechanism learns\cite{kim2021how}.
Recently, there has been an increasing amount of research addressing the expressive power of the graph attention mechanism \cite{GATv2}\cite{Balcilar.2021} and the generation of interpretable explanations for predictions\cite{Ying.2019}. Knyazev \textit{et al.} \cite{understandGAT} showed for special graph-pooling based attention algorithms \cite{Ying.2018}\cite{Gao.2018} that under typical conditions the effect of the attention mechanism is negligible and that performance is very sensitive to initialization of the attention model. By using simulated data, the authors analyzed performance and the assigned attention scores for solving different graph classification tasks. In contrast to the analyzed approaches of \cite{understandGAT}, currently popular GAT realizations use the softmax function to compute attention scores. The softmax is an approximation to the $\arg \max$ function. The advantage of using softmax is that the function is continuous and differentiable, which allows a more stable training characteristic \cite{Goodfellow.2016}.
A similar result on the relevance of input-specific attention using sequential data was proposed in \cite{Hassid.07.11.2022}. In their work, the authors analyzed the importance of the attention assignment in pre-trained language models and showed that the attention mechanism is not as important as typically thought. By replacing the input-dependent attention mechanism with a constant attention matrix, all models still achieved competitive performance. Since attention mechanisms weigh the representation of different inputs, it is often assumed that attention scores can be used for interpreting a model and identify causal relations. Serrano \textit{et al.} \cite{Serrano.2019}, however, showed that the scores do not necessarily indicate importance. Contrarily to these findings, Renz \textit{et al.} \cite{Renz.25.10.2022} proposed a transformer-based approach for path planning, where the most relevant objects for the decision are identified by using the assigned attention scores on the object level. Their results show, that their approach can focus on the most relevant object in the scene, even when it is not geometrically close. Hence, there is still a big dissent about the interpretability of assigned attention scores. 

In this work, the interpretation of attention scores of GATs as relevance indicator is investigated. Knowledge about which relations mainly influenced a networks prediction provides valuable insights.
\begin{figure}[t!]
	\centering
	\vspace{7pt}
	\includegraphics[width=0.6\columnwidth]{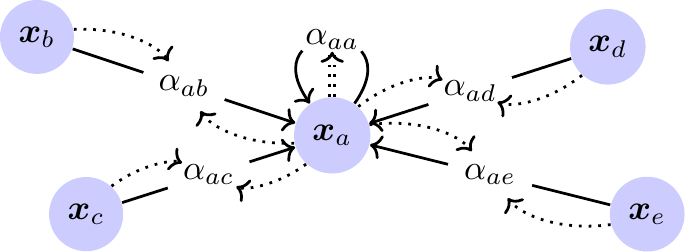}
	\caption{Concept of attentional message passing on graphs.}
	\label{fig:GAT}
	\vspace{\vneg}
\end{figure}
\section{Preliminaries}
\textbf{Graph definition}: Let $\mathcal{G} = (\mathcal{V}, \mathcal{E})$ be a graph composed of nodes $\mathcal{V}=\{1, \dots, n\}$ and edges $\mathcal{E} \subseteq \mathcal{V} \times \mathcal{V}$. An edge from a node $j$ to a node $i$ is represented by $(i,j) \in \mathcal{E}$. If all edges are bidirectional the graph is denoted as undirected; and directed if otherwise. The graph $\mathcal{G}$ can be represented through its adjacency matrix $\bm{A} = \{0, 1\}^{|\mathcal{V}|\times|\mathcal{V}|}$. If the graph is weighted, an additional weight matrix $\bm{W}~\in~\mathbb{R}^{|\mathcal{V}|\times|\mathcal{V}|}$, indicating the weighting of each edge, can be defined.

\textbf{Graph Neural Networks}: A GNN is based on the concept of message-passing. Each node $i \in \mathcal{V}$ of the graph structure is assigned an initial feature $\bm{h}_i^{(0)} \in \mathbb{R}^d$. The network updates the node features through successively applying an $\textit{AGGREGATE}$ and $\textit{UPDATE}$ step: 
\begin{align}
\bm{c}_i^{(k)} &= \text{AGGREGATE}^{(k)} \left( \{ \bm{h}_j^{(k-1)}: j \in \mathcal{N}_i \} \right)\\
\bm{h}_i^{(k)} &= \text{UPDATE}^{(k)} \left(\bm{h}_i^{(k-1)}, \bm{c}_i^{(k)} \right).
\end{align}
In the aggregation step of node $i$ the network aggregates the features of its neighboring nodes $j \in \mathcal{N}_i $ into the latent variable $\bm{c}_i^{(k)}$. In the update step the information of the neighboring nodes $\bm{c}_i^{(k)}$ is combined with the current features of node $i$ to update its features. One repetition of these two steps can be regarded as one layer $k = \{1, \dots, K\}$. Different types of GNNs differ in the definition of the aggregation and update function.

\textbf{Graph Attention Networks}: In GATs, the features of the neighboring nodes are aggregated by computing attention scores $\alpha_{ij}$ for every edge~$\left(i,j\right)$. As shown in Fig.~\ref{fig:GAT}, the idea is to perform a weighted sum over the neighboring nodes $j \in \mathcal{N}_i$, where the attention score $\alpha_{ij}$ indicates the importance of the features of neighbor~$j$ to the node~$i$. Note, that if node $i$ has a self-connection, it formally is considered as a neighboring node to itself.

There are different ways to compute attention scores. A very popular variant is the GATv2\cite{GATv2}, where the network updates the features of node $i$ as shown in \mbox{Eq. \ref{eq:Nscore}-\ref{eq:GATV2UPDATE}}. The equations correspond to the default implementation of GATv2\cite{GATv2} in the PyTorch Geometric framework~\cite{pytorchGATv2CONV}. 
\begin{align}
e(\tilde{\bm{h}}_i, \tilde{\bm{h}}_j) &= \bm{a}^\text{T}\text{LeakyReLU}\left({\boldsymbol{\Theta}}_R\tilde{\bm{h}}_i +{\boldsymbol{\Theta}}_L \tilde{\bm{h}}_j \right) \label{eq:Nscore}\\
\alpha_{ij} &= \text{softmax}_j(e (\bm{\tilde{h}}_i, \bm{\tilde{h}}_j) ) \label{eq:sftmax}\\
\bm{h}_i' &= \bm{b}  + \sum_{j \in \mathcal{N}_i} \alpha_{ij} {\boldsymbol{\Theta}}_L\tilde{\bm{h}}_j
\label{eq:GATV2UPDATE}
\end{align}
The attention scores are determined by the scoring function (Eq.~\ref{eq:Nscore}),
where $\bm{a}\in \mathbb{R}^{d'}$,  $\boldsymbol{\Theta}_p \in \mathbb{R}^{d' \times (d+1)}$ for \mbox{$p\in\{R,L\}$} are learned and $\tilde{\bm{h}}_q = [1, \bm{h}_q^\text{T}]^\text{T}$ for $q\in\{i,j\}$ are node representations. In Eq.~\ref{eq:sftmax}, the resulting values of the scoring function $e(\tilde{\bm{h}}_i, \tilde{\bm{h}}_j)$ are normalized using softmax such that $\sum_j \alpha_{ij} = 1$. The normalized attention scores are used to update the feature representation by computing a weighted sum as described in Eq. \ref{eq:GATV2UPDATE}, where $\bm{b} \in \mathbb{R}^{d'}$ is a learnable parameter. Since this work focuses on one-layered realizations, layer indexing is neglected and the updated node representation is denoted as $\bm{h}_i'$. However, if several layers are used, the updated feature representation $\bm{h}_i'$ is usually passed through an additional activation function \cite{GATv2}. Initially, each node representation is parameterized based on the corresponding data features \mbox{$\bm{h}_q = \bm{x}_q$}. 

In comparison to the originally proposed GAT~\cite{Velickovic.30.10.2017}, GATv2\cite{GATv2} allows dynamic attention assignment. The static form of GAT is only able to perform a limited kind of attention, where the ranking of the attention scores is unconditional on the query node\cite{GATv2}. The attention mechanism introduced by \cite{Vaswani.2017} has been shown to be strictly weaker than GATv2 \cite{GATv2}.
Therefore, the focus of this paper lies in the analyzation of GATv2. 

\section{Pitfalls of GATv2 and Architectural Optimization}
\label{sec:pitfalls}
In the following, potential pitfalls of GATv2 are identified and resulting limitations are elaborated. Additionally, architectural optimizations to surpass the shortcomings are proposed.
The following subsections identify three key factors that limit the expressive power and interpretability of GATv2. While subsections \textit{A} and \textit{B} are mainly relevant for automotive applications, subsection \textit{C} reveals a general pitfall.

For a better understanding, the aspects are supported by an automotive example: Let $\mathcal{G}$ be an unweighted graph, where nodes represent participants of a traffic scenario and edges indicate inter-dependencies. In particular, vehicle (node) $i$ and the nearby traffic participants $j \in \mathcal{N}_i$ are involved in an intersection scenario. Each node feature holds the position and velocity of the respective participant. Graph $\mathcal{G}$ contains self-connections, such that features of the query node~$i$ are considered in the update function Eq.~\ref{eq:GATV2UPDATE}. On basis of the scenario context $\mathcal{G}$, GATv2 is used to extract relevant scenario information and enrich graph node representations for downstream tasks like future motion prediction. Due to the attention mechanism, GATv2 dynamically assigns weights to each relation. For an interpretable feature embedding, relations to especially relevant neighboring traffic participants should be assigned high attention scores.

 \subsection{Attention assignment for query node}
 In certain scenarios, the future motion of a vehicle (query node)~$i$ might depend equally on its own motion and one particular traffic participant (node)~$j$. Under the assumption that attention scores indicate relevance, in such scenarios, identical attention scores should be assigned to the corresponding query node $i$ and neighboring node~$j$.
 While equal attention scores $\alpha_{ii}=\alpha_{ij}$ obviously result if $\bm{h}_i = \bm{h}_j$, this is a very particular case. In the introduced graph setup $\mathcal{G}$, for example, two participants cannot have the exactly same features, as this would mean they are at the same position. Nevertheless, since the attention scores are computed based on non-injective operations, the same attention scores $\alpha_{ii}=\alpha_{ij}$ for node~$i$ and $j$ can theoretically be induced even if $\bm{h}_i \neq \bm{h}_j$. In practice, however, this might be challenging to learn and require additional model complexity. In applications where high relevance for the query node $i$ is apparent, including it into the attention sharing introduces avertible complexity. By assigning a fixed attention score, e.\,g. $\alpha_{ii}=1$, to the query node and performing the attention assignment exclusively based on the neighboring nodes $j\in \mathcal{N}_i \setminus i$, the optimization constraints are relaxed and, hence, learning can be eased.
 \subsection{Subtraction not possible}
 The update function Eq.~\ref{eq:GATV2UPDATE} restricts the updated feature representation $\bm{h}_i'$ to be a weighted sum of all neighboring nodes. Since all neighboring nodes $j \in \mathcal{N}_i$, potentially including node~$i$, are transformed by the same matrix $\boldsymbol{\Theta}_L$, no feature subtraction is possible. In the automotive domain, however, important information often arises from feature differences, e.\,g., distance between traffic participants or relative velocities. The weighted sum of the participant's position and/or velocity is rather inconclusive. Hence, different transformations for the features of query node~$i$ and the neighboring nodes~$j$ in the update function are reasonable. By taking this and the subsection \textit{A} into account, the update function Eq.~\ref{eq:GATV2UPDATE} of GATv2 is adapted to
 \begin{equation}
 \bm{h}_i' =\bm{b} +  {\boldsymbol{\Theta}}_n\tilde{\bm{h}}_i + \sum_{j \in \mathcal{N}_i, i \neq j} \alpha_{ij} {\boldsymbol{\Theta}}_L\tilde{\bm{h}}_j,
 \label{eq:GATupdateAB}
 \end{equation}
 where $\boldsymbol{\Theta}_n\!\in\!\mathbb{R}^{d' \times (d+1)}$ is a learnable parameter. The introduced parameter $\boldsymbol{\Theta}_n$ enables to compute subtractions of the neighboring node's features and the query node $i$. As suggested in subsection \textit{A}, the attention value defaults to $\alpha_{ii}=1$. In the following, GAT variants that update the node features using Eq.~\ref{eq:GATupdateAB} are referred to as \mbox{GAT-$\Theta_n$}.
\subsection{Gradient for $\boldsymbol{\Theta}_{R}$ sensitive to initialization}
\renewcommand*{\thefootnote}{\fnsymbol{footnote}}
\renewcommand\footnoterule{\rule{0.4\linewidth}{0.8pt}}
As shown in this subsection, the performance of GATv2 is highly sensitive to network initialization.
%The gradient for the parameter $\theta_{R}$ is highly sensitive to initialization.
Due to the model definition of GATv2, the gradient for the parameter $\boldsymbol{\Theta}_{R}$ is prone to become zero; especially in sparse and simple graph structures. A zero gradient effects that the values of $\boldsymbol{\Theta}_{R}$ are not adopted during the optimization step. While this characteristic is desired if the parameter is close to an optimal parameterization, a zero gradient due to initialization and/or model design is unintended. As a consequence of GATv2 architecture proposed in \cite{GATv2} (as well as \cite{Velickovic.30.10.2017}), the parameter $\boldsymbol{\Theta}_{R}$ can potentially stay unchanged during a complete training or its optimization can stagnate early on in the training process although parameterization is not optimal. If the parameter $\boldsymbol{\Theta}_{R}$ is not learned properly, the underlying attention mechanism assigns attention scores unconditioned on the query node~$i$ and expressiveness is limited. 

Note, that the gradient always depends on the loss \mbox{function $\mathcal{L}$}. The following elaboration holds for the common case, that the loss is defined as the final prediction error and no intermediate loss is introduced.

To prove the proposition that the learning process is sensitive to initialization, the gradient ${}^\text{B}\boldsymbol{\Theta}_{R}$ for the parameter \mbox{$\boldsymbol{\Theta}_{R}= [\boldsymbol{\Theta}_{R_b}, \boldsymbol{\Theta}_{R_w}]$}, composed of the weight~$\boldsymbol{\Theta}_{R_w}$ and bias~$\boldsymbol{\Theta}_{R_b}$, is computed. Based on the chain rule of calculus, the gradients w.\,r.\,t. the model's parameters can be determined.\footnote{The complete derivation of the gradients is published at https://mariiilyn.github.io/2023/04/22/Gradient-Derivation-in-Graph-Attention-Networks.html.}
In the one-layered GATv2, the gradient ${}^\text{B}\boldsymbol{\Theta}_{R}$ w.\,r.\,t. the node~$i$ of feature dimension $d=1$, such that $\boldsymbol{\Theta}_R \in \mathbb{R}^{d'\times2}$, is defined as
\begin{equation}
{}^\text{B}\Theta_{R_w, i}^{t} = \frac{\partial \mathcal{L}}{\partial h_i'}\!\cdot\!
\left[
h_i a^t\sum_{j,k}^{\mathcal{S}_i} \alpha_{ij}\alpha_{ik} (A_j\! - \!A_k)(s_{ij}^t-s_{ik}^t)
\right]\!
\label{eq:dev_GATweight}
\end{equation}
and
\begin{equation}
{}^\text{B}\Theta_{R_b, i}^{t} = \frac{\partial \mathcal{L}}{\partial h_i'}\!\cdot\!
\left[ 
a^t\sum_{j,k}^{\mathcal{S}_i} \alpha_{ij}\alpha_{ik} (A_j \!- \!A_k)(s_{ij}^t-s_{ik}^t)
\right]\!,
\label{eq:dev_GATbias}
\end{equation}
where superscript $t = \{1, \dots, d'\}$ indicates the selected component of a vector representation, $A_q = \sum_t (\bm{\Theta}_{L} \tilde{\bm{h}}_q)^t $ for $q\in\{j,k\}$ and $\mathcal{S}_i=[\mathcal{N}_i]^2$ is the set of all the subsets of $\mathcal{N}_i$ with exactly two elements and no pairing repetitions. The parameter $s_{ij}$ is the derivative of the LeakyRELU activation.

In GATv2 the derivative of the LeakyRELU is
\begin{equation}
\frac{\partial \text{LeakyRELU}(x_{ij})}{\partial x_{ij}} = 
\begin{cases}
s_{ij}^t=1, & \text{if}\ x_{ij}^t>0 \\
s_{ij}^t=s_n, & \text{if}\ x_{ij}^t<0 
\end{cases},
\end{equation}
where $x_{ij}^t = (\boldsymbol{\Theta}_R\tilde{\bm{h}}_i +{\boldsymbol{\Theta}}_L \tilde{\bm{h}}_j )^t$ and $s_n$ is the slope of the LeakyRELU for negative values of $x_{ij}^t$.\\
While there are various parameterizations such that 
$
{}^\text{B}\Theta_{R_w, i}^{t}=
{}^\text{B}\Theta_{R_b, i}^{t} = 0,
$
the gradient ${}^\text{B}\boldsymbol{\Theta}_{R}^{t}$ is especially likely to become zero due to the term $(s_{ij}^t  - s_{ik}^t)$. For the gradient $\forall i \in \mathcal{V}:{}^\text{B}{\Theta}_{R, i}^{t}$ it holds that if
\begin{align}
\text{sign}({x}_{ij}^t) &= \text{sign}({x}_{ik}^t) \quad \forall \{j,k\} \in \mathcal{S}_i \\ \implies  &{}^\text{B}{\Theta}_{R, i}^t=0.
\end{align}
If the intermediate representations $x_{ij}^t$ and $x_{ik}^t$ for all pairings $\{j,k\} \in \mathcal{S}_i $ have the same sign, the component $t$ of the gradient ${}^\text{B}\bm{\Theta}_{R,i}$ becomes zero. The derivatives Eq.~\ref{eq:dev_GATweight} and Eq.~\ref{eq:dev_GATbias} show, that this characteristic is not limited to certain graph or experimental setups, but is a general pitfall of GATv2. The occurrence of this condition strongly depends on the data and the parameter initialization.
Due to the concept of shared weights, the overall gradient for updating $\bm{\Theta}_{R}$ is 
\begin{equation}
{}^\text{B}\bm{\Theta}_{R}= \sum_i^{|\mathcal{V}|}	{}^\text{B}\bm{\Theta}_{R,i}.
\end{equation}
Hence, the gradient $ {}^\text{B}\bm{\Theta}_{R}$ is generally less likely to become zero in dense graphs or graphs that consist of many nodes. In fact, however, most of real-world data structures are sparse \cite{Danisch.2018}\cite{Bhadra.2009}. Also in automotive applications, where often star graphs are used to represent traffic scenarios, graphs are usually sparse and small.
If and only if 
$\exists~\{j,k\}~\in~\mathcal{S}_i~\ni~\text{sign}({x}_{ij}^t)~\neq~\text{sign}({x}_{ik}^t)$ the gradient ${}^\text{B}{\Theta}^t_{R,i} \neq 0$ is considered in optimizing the network parameters. If the majority of the pairings (or none in the worst-case) don't meet this requirement, the training process might fail to optimize $\bm{\Theta}_{R}$. Ultimately the attention score assignment becomes unconditioned on the query node $i$. This leads to the assumption that only experiments, which are based on an attention score assignment independent of the query node $i$ can be resolved reliably. In the following, two possible solutions to bypass this problem are elaborated.
\subsubsection{Alternative activation function}
The problem arises from the fact, that the derivative of the LeakyRELU results in one of two possible values. By replacing the LeakyRELU with an alternative activation function the problem can be circumvented. An ideal activation function meets the following two requirements.\\ 
\textbf{Injective Derivative} The derivative of the activation function should be an injective function such that $\forall~a,b~\in~X,\,f(a)~=~f(b)~\implies~a =b$. An injective derivative prevents a zero gradient due to the term $(s_{ij}^t - s_{ik}^t)$.\\
\textbf{Non-saturating} To support the selective nature of the attention mechanism, the activation function should not be saturating, hence, it should satisfy $\lim_{|x|\rightarrow\infty} |\nabla f(x)| \neq 0$. 

One conceivable activation function, which partly satisfies both requirements is the softplus activation function
\begin{equation}
\text{softplus}(x)= \ln \left(1+e^x\right).
\end{equation}
In the following, if the LeakyRELU in the aggregation function Eq.~\ref{eq:Nscore} is replaced by the softplus, the GAT realization is denoted with the superscript~${}^+$, e.\,g., GAT-${\Theta}_n^+$. 
\subsubsection{Adapt update function}
Instead of introducing a new parameter $\boldsymbol{\Theta}_n$ in the update function as done in GAT-${\Theta}_n$, $\boldsymbol{\Theta}_R$ can be used to transform the query node such that
\begin{align}
\bm{h}_i' &= \bm{b} + {\boldsymbol{\Theta}}_R\tilde{\bm{h}}_i + \sum_{j \in \mathcal{N}_i, i \neq j} \alpha_{ij} {\boldsymbol{\Theta}}_j\tilde{\bm{h}}_j.
\label{eq:GATwi}
\end{align}
By doing this, the gradient ${}^\text{B}\boldsymbol{\Theta}_{R}$ gets an additional gradient component which arises from $\frac{\partial (\boldsymbol{\Theta}_R\tilde{\bm{h}}_i)}{\partial\boldsymbol{\Theta}_R}$ in the update function and is only zero if $\tilde{\bm{h}}_i=\bm{0}$. In fact, using parameter $\boldsymbol{\Theta}_{R}$ in the update function is reasonable since this maps the node features into the same representation space that the attention score assignment is based on. GAT realizations that update the node features using Eq.~\ref{eq:GATwi} are referred to as GAT-$\Theta_R$. 
\section{Dataset and Experiments}
\label{sec:experiments}
In order to evaluate the expressive power and interpretability of attention scores, a synthetically generated dataset $\mathcal{D}$ is introduced. Real world datasets hardly provide ground truth attention scores which are required to assess the attention mechanism. The controlled environment of the synthetic dataset, however, enables access to accurate ground truth attention values. On basis of the dataset $\mathcal{D}$, node-level regression tasks should be solved using GATv2. Throughout the experiments, the graph architecture $\mathcal{G} =(\mathcal{V}, \mathcal{E})$ is not varied. The graph is specified as star graph, such that each node has a bidirectional connection to one central node but no connection to any other node. Additionally, the central node is provided a self-connection. In Fig. \ref{fig:GAT}, the experimental star graph architecture is illustrated.

\begin{figure}[t!]
	\begin{subfigure}[b!]{0.49\columnwidth}
	\centering
	\scalebox{.7}{ 	\large
		\begin{tikzpicture}
		\begin{axis}[
		scaled ticks=false,
		xmin=0,
		xmax = 0.5*pi,
		ymin= 0,
		ymax = 1.05,
		width=6cm,
		height=4.0cm,
		xlabel=(a) $\mathcal{X}_\text{I}$,
		ylabel=relevance,
		]
		\addplot[domain=0:0.5*pi, red, ultra thick,smooth] {(sin(deg(x))};
		\end{axis}
		\end{tikzpicture}
	}
	\caption*{ }
	\label{fig:rel_LIN}
\end{subfigure} \hfill
	\begin{subfigure}[b!]{0.49\columnwidth}
	\centering
	\scalebox{.7}{ 	\large
	\begin{tikzpicture}
	\begin{axis}[
	scaled ticks=false,
	xmin=0,
	xmax = pi,
	ymin= 0,
	ymax = 1.05,
	width=6cm,
	height=4.0cm,
	xlabel=(b) $\mathcal{X}_{\text{II}}$,
	]
	\addplot[domain=0:pi, red, ultra thick,smooth] {(sin(deg(x))};
	\end{axis}
	\end{tikzpicture}
	}
	\caption*{}
	\label{fig:rel_SIN}
	\end{subfigure}
	\vspace{-0.8cm}
	\caption{Relevance function used for simulation, where (a) represents the strictly monotonic relevance function and (b) the parabolic relevance function.}
	\label{fig:experiments}
	\vspace{\vneg}
\end{figure}
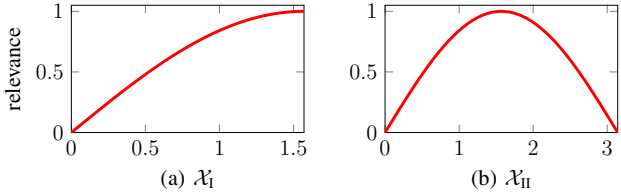

The $m$-th sample of the dataset is defined as \mbox{$\mathcal{D}_m:= \left(\mathbf{X}_m,~\bm{Y}_m,~\bm{\mathcal{A}}_m \right)$}, where $\mathbf{X}_m~=~\{\bm{x}_1^m,~\dots,~\bm{x}_{|\mathcal{V}|}^m\}^\text{T}$ contains the input features $\bm{x}_i\in\mathbb{R}^{d}$ for each node $ i \in \mathcal{V}$ and \mbox{$\bm{\mathcal{A}}_m\in \{1,0\}^{|\mathcal{V}|\times |\mathcal{V}|}$ }the true attention scores. \mbox{$\bm{Y}_m=\{\bm{y}_1^m,~\dots,~\bm{y}_{|\mathcal{V}|}^m\}^\text{T}$} contains the ground truth for the node-level regression task, where $\bm{y}_i^m\in\mathbb{R}^d$ is the ground truth for node $i$. 
As formulated in Eq.~\ref{eq:selectTrue} and Eq.~\ref{eq:taskTrue}, $\bm{y}_i^m$ is the feature difference of the query node $i$ and its most relevant neighboring node $j$. The relevance of the node pairings are simulated by the relevance function $r(\bm{x}_i, \bm{x}_j)$. 
\begin{align}
\bm{x}_r^m &= \arg \max_{j \in \mathcal{N}_i} r(\bm{x}_i^m, \bm{x}_j^m) \label{eq:selectTrue}\\
\bm{y}_i^m &= \bm{x}_r^m - \bm{x}_i^m. \label{eq:taskTrue}
\end{align}
In the individual experiments, the complexity of the simulated relevance function $r(\bm{x}_i, \bm{x}_j)$ is varied. As indicated in Eq.~\ref{eq:selectTrue}, all experiments are defined such that only one node out of all neighboring nodes is to be selected for solving the regression task correctly. Hence, the ground truth attention for node $i$ is an one-hot encoded vector $\bm{\alpha}^m_i := \text{col}_i \bm{\mathcal{A}}_m$ with 1 at the index of the relevant node and 0 everywhere else. The simulation is designed such that two nodes are never equally relevant.
The basis of the simulated relevance assignment is the sinus function, where
\begin{equation}
r(\bm{x}_i^m, \bm{x}_j^m) = \sin(\bm{x}_j^m - \bm{x}_i^m).
\end{equation}
\begin{figure}
	\centering
	\includegraphics[width=0.8\columnwidth]{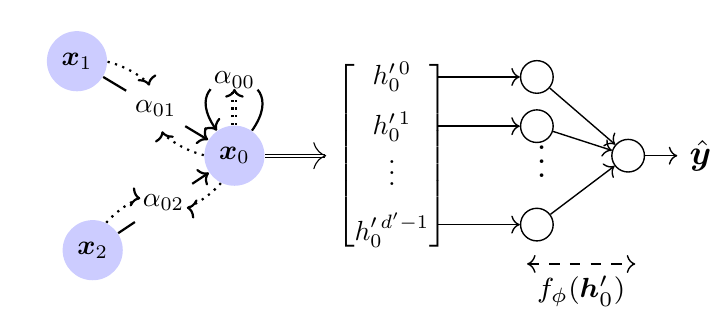}
	\caption{Exemplary graph setup during the experiment with $\mathcal{N}_{0} = \{0, 1,2\}$ and the additional feed-forward layer $f_\phi$ which is required only if $d' \neq d$.}
	\label{fig:ModelSetup} 
	\vspace{\vneg} 
\end{figure}
For each data sample $m$, the values $\forall i \in \mathcal{V}: \bm{x}_i$ are randomly sampled from the value range $\mathcal{X}$. Through adaption of the value range $\mathcal{X}$ different concepts of relevance are introduced (c.\,f. Fig.~\ref{fig:experiments}). The experiments of this work, evaluate the performance in the node-level regression task based on two different value ranges $\mathcal{X}$, i.\,e., complexities of relevance functions. For better traceability, the experiments are intentionally kept simple and the input feature dimension is defined as scalar $d=1$ such that $x_i^m, y_i^m \in \mathbb{R}$. For the central node, which is designated as $i=0$, it holds that $x_j^m-x_0^m>=0$.
Due to the star architecture of the graph only the central node has multiple neighbors. All other nodes only have the central node as neighbor and would inherently always select the correct node. Without loss of generality, training and evaluation is therefore only based on the central node. Since GATs are based on the concept of shared weights, the learned logic is scalable to more complex graph structures. \\ 
An one-layered GAT $g_{\boldsymbol{\Theta}}(\mathbf{X}_m)$ is intended to adapt its model weights $\boldsymbol{\Theta}$ during the training process to approximate the correct mapping of Eq.~\ref{eq:selectTrue}-\ref{eq:taskTrue}. Each model is given enough degrees of freedom to theoretically be able to solve the task. However, an one-layered GAT (Eq.~\ref{eq:Nscore}-\ref{eq:GATV2UPDATE}) inevitably faces a dimension misfit if the dimension of the regression task ground truth $d=1$ is unequal to the latent dimension $d'$ required to solve the task. To bypass this problem, an additional learnable feed-forward layer $f_\phi:~\mathbb{R}^{d'}~\rightarrow~\mathbb{R}^{d}$ is added to the node-level output of the GAT if $d' \neq d$. As indicated in Fig.~\ref{fig:ModelSetup}, the feed-forward layer $f_\phi$ maps the updated feature of node $i$ to the regression task dimension. Since the feed-forward layer $f_\phi$ gets only the updated node feature $\bm{h}_i'$, the correct node selection and feature aggregation has to be learned in the GAT model. 

\textbf{Experiment I: Strictly Monotonic Relevance Function}\\
By restricting the value range to $\mathcal{X}_\text{I} \in [0; 0.5\pi]$, the resulting relevance function is strictly monotonic, such that $a~<~b~ \rightarrow~r(a)~<~r(b)$. In Fig. \ref{fig:experiments}~(a) the relevance function over the defined value range is plotted. In this experiment the required latent dimension to solve the selection task is $d'=1$. Therefore, no additional learnable layer is required. The prediction $\hat{y}_i^m$ is the updated feature of node $i$:
\begin{equation}
	\hat{y}^m_i =g_{\boldsymbol{\Theta}}(\mathbf{X}_m)_i.
\end{equation} 
In the automotive application example, this would equal the task of selecting the most relevant neighboring vehicle $j$ for vehicle $i$ based on their relative velocity: The higher the relative velocity, the higher the importance assigned to node $j$ by the simulation. The regression task is to output their relative velocity in the updated node feature $h_i':= g_{\boldsymbol{\Theta}}(\mathbf{X}_m)_i$.
 
\textbf{Experiment II: Parabolic Relevance Function}\\
By setting the value range to $\mathcal{X}_{\text{II}} \in [0; \pi]$ a parabolic relevance function is achieved, which is slightly more complex than the strictly monotonic function. Fig. \ref{fig:experiments}~(b) shows the resulting relevance function over the defined value range. For the GAT to be able to approximate the relevance function, the latent dimension has to be set to $d'=2$. Since this definition causes a dimension misfit in the regression task, an additional learnable layer $f_{\phi}$ is introduced for this experiment such that
\begin{equation}
\hat{y}^m_i = f_{\phi} \left(g_{\boldsymbol{\Theta}}(\mathbf{X}_m)_i\right).
\end{equation}
A LeakyRELU is applied at the beginning of layer $f_{\phi}$.

In the automotive application example, this would equal the task of selecting the closest neighboring vehicle $j\in\mathcal{N}_i$, which the vehicle $i$ can react on, and compute their distance.\\ 

In the experiments, graph $\mathcal{G}$ consists of $|\mathcal{V}| = 3$ nodes in total, with node $i=0$ being the central node and $j~\in~\mathcal{N}_0 =~\{0,1,2\}$ its neighboring nodes. For training as well as for testing $M=20\,000$ data samples are generated. Each model implementation is trained with unsupervised attention only, using the loss function
\begin{equation}
\mathcal{L}(\hat{y}^m, y^m) =  \hat{y}_i^m -y_i^m.
\end{equation}

\section{Evaluation}
% axis style, ticks, etc
\pgfplotsset{interpret/.append style={
		grid=major,
		grid style={dashed,gray!30},
		x label style={at={(axis description cs:0.5,-0.12)}, anchor=north},
		y label style={at={(axis description cs:-0.25,.5)}, anchor=south},
		xlabel={\small{Attention $\alpha_{ij}$}},          % default put x on x-axis
%		ylabel={Percentage $[\%]$},          % default put y on y-axis
		label style={font=\small},
		title style={yshift=-8pt, font=\small},
		tick label style={font=\scriptsize}  
}}

\def \bwidth {5pt}
\def \offset {2.5cm}
\def \size {3.8cm}
\def \nheight {3.6cm}
\def \neg {-0.5cm}
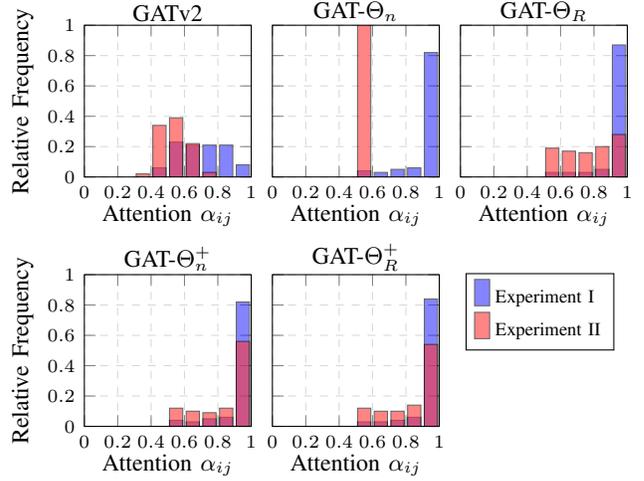
\begin{figure}[t!]
	\vspace{-2pt}
	\begin{subfigure}[]{\columnwidth}
		\begin{tikzpicture}
		\begin{axis}[
		interpret,
		title={GATv2},
		ymin=0,
		xmin=0,
		xmax=1,
		ymax=1,
		ylabel={\small{Relative Frequency}},
width = \size,
height = \nheight
		]
		% GATv2 - Linear
		\addplot[ybar,bar width=\bwidth,fill=blue!80,opacity=0.6] 
		coordinates {(0.45,0.06)(0.55,0.23)(0.65,0.21)(0.75,0.21)(0.85,0.21)(0.95,0.08)};
		\addplot[ybar,bar width=\bwidth,fill=red!80,opacity=0.6] 		
		coordinates {(0.35,0.02)(0.45,0.34)(0.55,0.39)(0.65,0.22)(0.75,0.03)};
		\end{axis}
\begin{axis}[
interpret,
xshift = \offset,
title={GAT-$\Theta_n$},
yticklabels={,,},
ymin=0,
xmin=0,
xmax=1,
ymax=1,
width = \size,
height =\nheight
]
\addplot[ybar,bar width=\bwidth,fill=blue!80,opacity=0.6]  
coordinates{(0.55,0.04)
	(0.65,0.03)
	(0.75,0.05)
	(0.85,0.06)
	(0.95,0.82)};
	\addplot[ybar,bar width=\bwidth,fill=red!80,opacity=0.6]  
	coordinates{(0.55,1)};
%	coordinates{(0.55,0.12)
%	(0.65,0.13)
%	(0.75,0.14)
%	(0.85,0.18)
%	(0.95,0.43)};
\end{axis}
\begin{axis}[
	interpret,
	xshift = 2*\offset,
	title={GAT-$\Theta_R$},
	yticklabels={,,},
	ymin=0,
	xmin=0,
	xmax=1,
	ymax=1,
width =\size,
height = \nheight
	]
	\addplot[ybar,bar width=\bwidth,fill=blue!80,opacity=0.6] 
	coordinates{(0.55,0.03)
		(0.65,0.03)
		(0.75,0.03)
		(0.85,0.05)
		(0.95,0.87)
	};
	\addplot[ybar,bar width=\bwidth,fill=red!80,opacity=0.6] 
	coordinates{(0.55,0.19)
		(0.65,0.17)
		(0.75,0.16)
		(0.85,0.20)
		(0.95,0.28)
	};
	\end{axis}
	\end{tikzpicture}
	\caption*{ }
	\vspace{\neg}
\end{subfigure} 
\newline
\begin{subfigure}[]{\columnwidth}
	\centering
	\vspace{-2pt}
	\begin{tikzpicture}
	\begin{axis}[
	interpret,
	width=\size,
	height=\nheight,
	ylabel={\small{Relative Frequency}},
	title={GAT-$\Theta_n^+$},
	ymin=0,
	xmin=0,
	xmax=1,
	ymax=1
	]
	\addplot[ybar,bar width=\bwidth,fill=blue!80,opacity=0.6] 
	coordinates{
		(0.55,0.04)
		(0.65,0.03)
		(0.75,0.05)
		(0.85,0.06)
		(0.95,0.82)
		};
	\addplot[ybar,bar width=\bwidth,fill=red!80,opacity=0.6] 
	coordinates{(0.55,0.12)
		(0.65,0.10)
		(0.75,0.09)
		(0.85,0.12)
		(0.95,0.56)
	};
	\end{axis}
\begin{axis}[
xshift = \offset,
interpret,
width=\size,
height=\nheight,
title={GAT-$\Theta_R^+$},
yticklabels={,,},
ymin=0,
xmin=0,
xmax=1,
ymax=1,
legend style={at={(1,0.5)},xshift=0.35cm,
	anchor=south west, nodes=right, minimum width=\bwidth},
legend columns=1,
legend image code/.code={
	\draw [#1] (0cm,-0.1cm) rectangle (0.2cm,0.25cm); },
legend entries={{\scriptsize Experiment I}, {\scriptsize Experiment II}},
]
\addplot[ybar,bar width=\bwidth,fill=blue!80,opacity=0.6] 
coordinates{(0.55,0.03)
	(0.65,0.03)
	(0.75,0.04)
	(0.85,0.06)
	(0.95,0.84)
};
\addplot[ybar,bar width=\bwidth,fill=red!80,opacity=0.6] 
coordinates{(0.55,0.12)
	(0.65,0.10)
	(0.75,0.10)
	(0.85,0.14)
	(0.95,0.54)
};
\end{axis}
\begin{axis}[
xshift = 2*\offset +0.5cm,
width=\size,
height=\nheight,
yticklabels={,,},
xticklabels={,,},
ticks=none,
axis line style={draw opacity=0},
ymin=0,
xmin=0,
xmax=1,
ymax=1,
]
\end{axis}
\end{tikzpicture}
\caption*{}
\end{subfigure}
\vspace{-18pt}
\caption{Confidence histograms for experiment I and experiment II. If a node is correctly identified as the most relevant neighboring node, the assigned attention score $\alpha_{ij}$ is considered as sample in the corresponding histogram bin.}
\label{fig:confidence}
%\vspace{\vneg}
\end{figure}

The dataset $\mathcal{D}$ is used for an empirical study of the performance and attention interpretability of one-layered, single-headed GAT realizations. In addition to GATv2, model variants based on the proposed architectural adaptions, namely GAT-$\Theta_R$, GAT-$\Theta_n$, GAT-$\Theta_R^+$ and GAT-$\Theta_n^+$, are evaluated.
Results of the different experiments are shown in Table~\ref{table:comparison}. 
\begin{table}[h!]
	\centering
	\begin{tabular}{cc|c|c|c|c|}
		&Architecture& TPR & ME $\pm \sigma^2$ & Max. Error   \\
		\hline
		\hline
		\parbox[t]{2mm}{\multirow{5}{*}{\rotatebox[origin=c]{90}{\scriptsize{\textit{Experiment I}}}}} 
		&GATv2	   	 & \textbf{1.0} & 0.029 $\pm$ 0.001 & 0.198  \\	
		&GAT-$\Theta_n$   & \textbf{1.0} & \textbf{0.000} $\pm$ \textbf{0.000 }&  {0.003}  \\
		&GAT-$\Theta_n^+$ & \textbf{1.0} & \textbf{0.000} $\pm$ \textbf{0.000} & \textbf{0.000}  \\ 
		& GAT-$\Theta_R$ & \textbf{1.0} & \textbf{0.000} $\pm$ \textbf{0.000} & \textbf{0.000}  \\ 
		&GAT-$\Theta_R^+$ & \textbf{1.0} & \textbf{0.000} $\pm$ \textbf{0.000} & \textbf{0.000}  \\ 
		\hline
		\hline
		\parbox[t]{2mm}{\multirow{5}{*}{\rotatebox[origin=c]{90}{\scriptsize{\textit{Experiment II}}}}} 
		&GATv2	   	 & 0.497 & 0.135 $\pm$ 0.036 & 1.658  \\	
		&GAT-$\Theta_n$ 	 & 0.497 & 0.105 $\pm$ 0.020 & \textbf{1.079} \\
		&GAT-$\Theta_n^+$  	 & \textbf{0.982} &\textbf{0.018} $\pm$ \textbf{0.004} & 1.677  \\
		&GAT-$\Theta_R$ & 0.937 & 0.039 $\pm$ 0.008 & 1.395\\ 
		&GAT-$\Theta_R^+$ &  0.964 & 0.020 $\pm$\ 0.006& 2.123\\
	\end{tabular}%
	\caption{Performance and selection accuracy based on the same model parameter initialization $e$.}
	\label{table:comparison}
	\vspace{\vneg}
\end{table}
The True Positive Rate (TPR)
\begin{equation}
\text{TPR} = \frac{\text{TP}}{\text{TP} + \text{FN}},
\label{eq:TPR}
\end{equation}
where TP is the number of true positives and FP of false negatives, indicates the percentage of test samples in which a node is correctly identified to be most relevant. If the highest attention score is assigned to the correct node it's considered a TP. The TPR is an unbiased estimator of the node selection accuracy.
Performance of the node-level regression task is evaluated trough the mean prediction error 
\begin{equation}
\text{ME} = \frac{1}{M} \sum_{m=1}^{M} (\hat{y}_i^m - y_i^m),
\label{eq:mse}
\end{equation}
the error variance $\sigma^2$ as well as the mean maximal error. 
\begin{figure*}
\begin{subfigure}[]{1.\textwidth}
\begin{tikzpicture}
\centering
\begin{axis}[
grid=major,
grid style={dashed,gray!30},
ylabel=ME ,
minor grid style={gray!25},
major grid style={gray!25},
height = 3.3cm,
xmin = 0,
xmax = 100,
width= \textwidth,
xticklabels={,,},
tick label style={/pgf/number format/fixed},
]
\addplot[line width=0.5pt,solid,color=blue, mark=x] table[line join=round, x= idx, y=mse,col sep=comma]{contentfiles/0GATnsftp.csv};
\addplot[line width=0.5pt,dashed,color=cyan, mark= star] table[x= idx, y=mse,col sep=comma]{contentfiles/0GATn.csv};
\addplot[line width=0.5pt,solid,color=black, mark=|] table[line join=round, x= idx, y=mse,col sep=comma]{contentfiles/0GATv2.csv};
\end{axis}
\begin{axis}[
yshift=-1.8cm,
xlabel=Initialization $e$, 
ylabel=TPR ,
grid=major,
grid style={dashed,gray!30},
minor grid style={gray!25},
major grid style={gray!25},
width= \textwidth,
height = 3.3cm,
xmin = 0,
xmax = 100,
x label style={at={(axis description cs:0.5,-0.25)}, anchor=north},
legend style={at={(1,0)},xshift=-0.1cm,yshift =0.1cm,
	anchor=south east, nodes=right, minimum size=0.4cm, inner sep=0pt},
legend columns=3,
legend entries={{\scriptsize GAT-$\Theta_n^+$},{\scriptsize GAT-$\Theta_n$},{\scriptsize GATv2},{\scriptsize GAT-$\Theta_R^+$},{\scriptsize GAT-$\Theta_n^+$}},
]
\addplot[line width=0.5pt,solid,color=blue, mark=x] table[line join=round, x= idx, y=tpr,col sep=comma]{contentfiles/0GATnsftp.csv};
\addplot[line width=0.5pt,dashed,color=cyan, mark= star] table[x= idx, y=tpr,col sep=comma]{contentfiles/0GATn.csv};
\addplot[line width=0.5pt,solid,color=black!80, mark=|] table[x= idx, y=tpr,col sep=comma]{contentfiles/0GATv2.csv};
\end{axis}
\end{tikzpicture}
\end{subfigure}
\vspace{-3pt}
\caption{Performance robustness for $E=100$ initializations.}
\label{fig:experimentsVAR}
\vspace{\vneg}
\end{figure*}
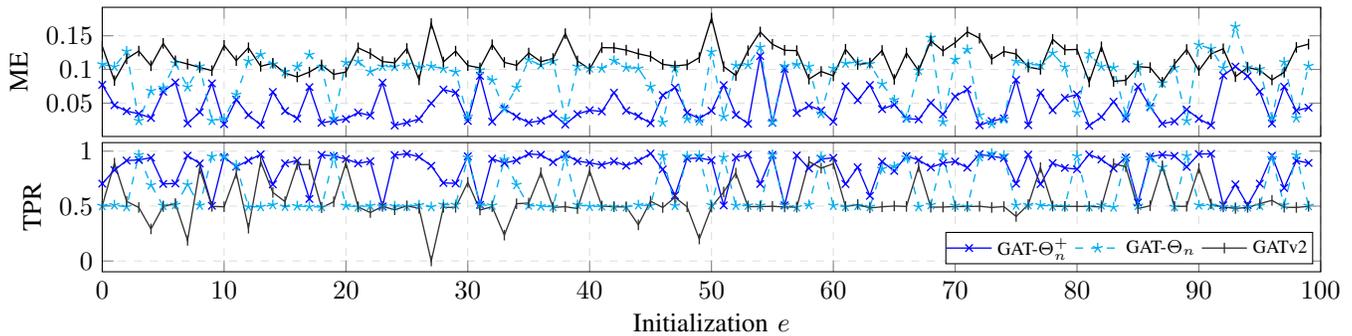

In experiment I, all GAT realizations perfectly manage to learn selecting the relevant neighboring node from all neighboring nodes $j \in \mathcal{N}_i$ with $\text{TPR}=1.0$. 
Due to the strictly monotonic nature of the relevance function, the attention assignment is substantially independent on the query node $i$. It holds that $\arg \max_j r(h_i, h_j)~=~\arg \max_j r(h_j)$ and subsequently, for selecting the relevant neighboring node the GAT parameter $\boldsymbol{\Theta}_{R}$ doesn't have to be optimized. Under those circumstances the proneness of its gradient ${}^\text{B}\boldsymbol{\Theta}_{R}$ to become zero does not negatively impact the performance. However, as the confidence histograms in Fig.~\ref{fig:confidence} indicate, GATv2 is strictly less selective than the models with proposed architectural adaptions. Results of the confidence histogram are used to evaluate if the attention values can be interpreted as a measure of node relevance. A perfect selection model would assign the value of $\alpha_{ij}=1.0$ to the most relevant node and zero to all others. The histogram of GATv2, however, suggests that GATv2 is non-selective and generally assigns comparatively low attention scores to the crucial node. Even though all other models assign an attention score between $0.9-1.0$ to the most relevant node in more than $\SI{80}{\%}$ of the samples, low attention scores still occur. It therefore has to be concluded that the highest attention values can suggest the most relevant node but assigned attention values are not equivalent to the corresponding degree of node relevance. Since GATv2 is unable to compute subtractions in the update function, the true mapping cannot be approximated properly. As a consequence GATv2, has a higher mean and maximal error in oppose to all other models. Note, that the limited expressivity of GATv2 update function enhances its non-selective characteristic.

\begin{figure}
		\vspace{4pt}
\begin{subfigure}[]{1.\columnwidth}
\begin{tikzpicture}
\centering
\begin{axis}[
boxplot/draw direction = y,
x axis line style = {opacity=0},
y axis line style = {opacity=60},
axis x line* = bottom,
axis y line = left,
enlarge y limits,
ymajorgrids,
xticklabels={,,}
xtick style = {draw=none}, % Hide tick line
tick label style={/pgf/number format/fixed},
ylabel = {ME},
ytick = {0, 0.05, 0.1, 0.15, 0.2},
boxplot/box extend=0.48,
width= \columnwidth,
height= 3.5cm,
]
\addplot+[boxplot,  fill=gray, draw=black] table[x= idx, y=mse,col sep=comma]{contentfiles/0GATv2.csv};
\addplot+[boxplot, fill=Maroon!60!orange, draw=black] table[x= idx, y=mse,col sep=comma]{contentfiles/0GATwi.csv};
\addplot+[boxplot, fill=cyan!60, draw=black] table[x= idx, y=mse,col sep=comma]{contentfiles/0GATn.csv};
\addplot+[boxplot, fill=orange!70, draw=black] table[x= idx, y=mse,col sep=comma]{contentfiles/0GATwisftp.csv};
\addplot+[boxplot, fill=NavyBlue!60, draw=black] table[x= idx, y=mse,col sep=comma]{contentfiles/0GATnsftp.csv};
\end{axis}
\end{tikzpicture}
\end{subfigure} 
\begin{subfigure}[]{1.\columnwidth}
	\begin{tikzpicture}
	\centering
	\begin{axis}[
	boxplot/draw direction = y,
	x axis line style = {opacity=0},
	y axis line style = {opacity=60},
	axis x line* = bottom,
	axis y line = left,
	enlarge y limits,
	ymajorgrids,
	xtick = {1, 2, 3, 4,5},
	xticklabel style = { anchor=center, yshift=-0.18cm, font=\small},%, rotate=60},
	xticklabels = {\scriptsize GATv2,\scriptsize GAT-$\Theta_{R}^{\,}$,\scriptsize GAT-$\Theta_{n}$, \scriptsize GAT-$\Theta_{R}^+$, \scriptsize GAT-$\Theta_{n}^+$},
	ylabel = {TPR},
	ytick = {0, 0.25, 0.5, 0.75, 1},
	boxplot/box extend=0.48,
	width= \columnwidth,
	height= 3.5cm,
	]
	\addplot+[boxplot,  fill=gray, draw=black] table[x= idx, y=tpr,col sep=comma]{contentfiles/0GATv2.csv};
	\addplot+[boxplot, fill=Maroon!60!orange, draw=black] table[x= idx, y=tpr,col sep=comma]{contentfiles/0GATwi.csv};
	\addplot+[boxplot, fill=cyan!60, draw=black] table[x= idx, y=tpr,col sep=comma]{contentfiles/0GATn.csv};
	\addplot+[boxplot, fill=orange!70, draw=black] table[x= idx, y=tpr,col sep=comma]{contentfiles/0GATwisftp.csv};
	\addplot+[boxplot, fill=NavyBlue!60, draw=black] table[x= idx, y=tpr,col sep=comma]{contentfiles/0GATnsftp.csv};
	\end{axis}
	\end{tikzpicture}
	\vspace{-4pt}
\end{subfigure}
\caption{Boxplots for the ME and TPR based on the results of $E=100$ random initializations.}
\label{fig:boxplot}
\vspace{\vneg}
\end{figure}
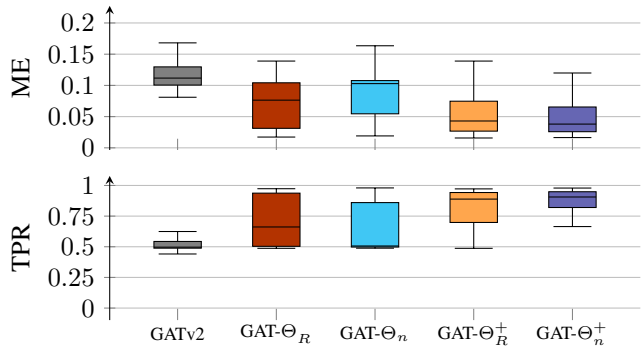

In experiment II, the relevance function is non-monotonic and, hence, query node information is crucial. Although the graph problem is only slightly more challenging, GATv2 performance drastically decreases. Only in about $\SI{50}{\percent}$ of the test samples the relevant node is identified correctly and regression task accuracy is significantly inferior. Due to a lack of gradient, the parameter $\boldsymbol{\Theta}_R$ isn't learned properly. Consequently, the attention assignment is optimized unconditioned on the query node parameter $\boldsymbol{\Theta}_R$. The hypothesis is supported by the fact, that by only introducing a new parameter in the update function as in \mbox{GAT-$\Theta_n$}, the TPR does not improve. While the extra parameter helps to improve regression task performance, it does not resolve the gradient issue of parameter $\boldsymbol{\Theta}_R$. The gradient ${}^\text{B}\boldsymbol{\Theta}_{R}$ is still prone to become zero. By additionally replacing the LeakyRELU as in GAT-$\Theta_n^+$, however, performance and TPR substantially improve. Also the adaptions made in GAT-$\Theta_R$, GAT-$\Theta_R^+$ benefit the selection accuracy as well as regression task performance. Yet, GAT-$\Theta_n^+$ outperforms all other GAT variants. When analyzing the confidence histogram in Fig.~\ref{fig:confidence} it can be seen that in experiment II the selective characteristics decreases for all models. This supports the presumption, that while the assigned attention score can be a useful measure of a node's significance, it may not always directly correlate with the node's overall importance within the graph context.
To show that the replacement of the LeakyRELU with the softplus makes the realizations strictly more robust to parameter initialization, the models are trained based on $E=100$ random parameter initializations. In Fig.~\ref{fig:experimentsVAR}, the ME and TPR of GATv2, GAT-$\Theta_n$ and GAT-$\Theta_n^+$ are evaluated for each initialization \mbox{$e \in \{0, \dots, E-1\}$}. The figure indicates, that only fortunate initializations enable the \mbox{GAT-$\Theta_n$} to be trained successfully, resulting in a good prediction performance, whereas the performance of \mbox{GAT-$\Theta_n^+$} is more robust. To avoid an overloaded visualization in Fig.~\ref{fig:experimentsVAR}, the models GAT-$\Theta_R$ and GAT-$\Theta_R^+$ are not considered. The boxplots in Fig.~\ref{fig:boxplot}, however, depicts the performance including spread and skewness for all GAT realizations. 

Both plots indicate, that GAT-$\Theta_n^+$ generally outperforms the other models in the regression task and is more accurate in selecting the relevant node. GATv2 is the least performant model. As can be seen in Fig.~\ref{fig:boxplot}, the variations GAT-$\Theta_n^+$ and GAT-$\Theta_R^+$, using the softplus activation, outperform their respective variants GAT-$\Theta_n$ and GAT-$\Theta_R$. Furthermore, the comparatively high performance variances of GAT-$\Theta_{n}$, GAT-$\Theta_{R}$ indicate an increased sensitivity to parameter initialization. Since \mbox{GAT-$\Theta_{n/R}$} and GAT-$\Theta_{n/R}^+$ only differ in their choice of activation function, it becomes clear that the LeakyRELU generally hinders the optimization process. 
 
\section{Conclusion}
In the automotive domain, GATs have become a popular approach to consider relational information within traffic scenarios for representation learning. 
This work identifies pitfalls of the established GATv2 that potentially impede the learning process. While real-world traffic scenarios are usually based on complex inter-dependencies, it is shown that GATv2 fails in solving already simple graph problems. Empirical evaluation based on different experiments evinces that the identified architectural pitfalls limit expressivity of GATv2 and model performance is highly sensitive to initialization. By adapting the model architecture as proposed in this work, limitations are removed and training is more robust to parameter initialization.
Additionally, the interpretability of assigned attention scores as measure of importance is evaluated. Results align with the findings of literature, that there is no general correlation between the assigned attention scores and node relevance. While attention scores can evidence relevance, it's not a reliable indication. Such architectural limitations should be considered very well to avoid false expectations.
The experiments of this work are based on synthetic and simple graph scenarios. The presented setup is appropriate for the elaboration of the fundamental pitfalls and limitations. In future works, the proposed model implementations will be tested and evaluated on real-world datasets.
 {
 	\bibliographystyle{IEEEtranS}
 	\bibliography{ref/ref_gat.bib}

% Generated by IEEEtranS.bst, version: 1.12 (2007/01/11)
\begin{thebibliography}{10}
\providecommand{\url}[1]{#1}
\csname url@samestyle\endcsname
\providecommand{\newblock}{\relax}
\providecommand{\bibinfo}[2]{#2}
\providecommand{\BIBentrySTDinterwordspacing}{\spaceskip=0pt\relax}
\providecommand{\BIBentryALTinterwordstretchfactor}{4}
\providecommand{\BIBentryALTinterwordspacing}{\spaceskip=\fontdimen2\font plus
\BIBentryALTinterwordstretchfactor\fontdimen3\font minus
  \fontdimen4\font\relax}
\providecommand{\BIBforeignlanguage}[2]{{%
\expandafter\ifx\csname l@#1\endcsname\relax
\typeout{** WARNING: IEEEtranS.bst: No hyphenation pattern has been}%
\typeout{** loaded for the language `#1'. Using the pattern for}%
\typeout{** the default language instead.}%
\else
\language=\csname l@#1\endcsname
\fi
#2}}
\providecommand{\BIBdecl}{\relax}
\BIBdecl

\bibitem{Balcilar.2021}
M.~Balcilar, G.~Renton, P.~H{\'e}roux, B.~Ga{\"u}z{\`e}re, S.~Adam, and
  P.~Honeine, ``Analyzing the expressive power of graph neural networks in a
  spectral perspective,'' in \emph{International Conference on Learning
  Representations}, 2021.

\bibitem{Bhadra.2009}
S.~Bhadra, C.~Bhattacharyya, N.~R. Chandra, and I.~S. Mian, ``A linear
  programming approach for estimating the structure of a sparse linear genetic
  network from transcript profiling data,'' \emph{Algorithms for molecular
  biology : AMB}, vol.~4, p.~5, 2009.

\bibitem{BruelGabrielsson.2020}
R.~{Br{\"u}el Gabrielsson}, ``Universal function approximation on graphs,'' in
  \emph{Advances in Neural Information Processing Systems}, {H. Larochelle},
  {M. Ranzato}, {R. Hadsell}, {M.F. Balcan}, and {H. Lin}, Eds., vol.~33, 2020,
  pp. 19\,762--19\,772.

\bibitem{Chen.2020}
D.~Chen, Y.~Lin, W.~Li, P.~Li, J.~Zhou, and X.~Sun, ``Measuring and relieving
  the over-smoothing problem for graph neural networks from the topological
  view,'' in \emph{Proceedings of the AAAI Conference on Artificial
  Intelligence}, 2020, vol.~34, pp. 3438--3445.

\bibitem{Danisch.2018}
M.~Danisch, O.~Balalau, and M.~Sozio, ``Listing k-cliques in sparse real-world
  graphs,'' in \emph{Proceedings of the 2018 World Wide Web Conference}, ser.
  WWW '18.\hskip 1em plus 0.5em minus 0.4em\relax Republic and Canton of
  Geneva, CHE: {International World Wide Web Conferences Steering Committee},
  2018, pp. 589--598.

\bibitem{Deac.2019}
A.~Deac, P.~Veli{\v{c}}kovi{\'c}, and P.~Sormanni, ``Attentive cross-modal
  paratope prediction,'' \emph{Journal of Computational Biology}, vol.~26,
  no.~6, pp. 536--545, 2019.

\bibitem{kim2021how}
{Dongkwan Kim} and {Alice Oh}, ``How to find your friendly neighborhood: Graph
  attention design with self-supervision,'' in \emph{International Conference
  on Learning Representations}, 2021.

\bibitem{Gao.2018}
H.~Gao and S.~Ji, ``Graph u-nets,'' in \emph{Proceedings of the 36 th
  International Conference on Machine (ICLM)}, 2018, vol.~36.

\bibitem{VectorNET.2020}
J.~Gao, C.~Sun, H.~Zhao, Y.~Shen, D.~Anguelov, C.~Li, and C.~Schmid,
  ``Vectornet: Encoding hd maps and agent dynamics from vectorized
  representation,'' in \emph{Computer Vision and Pattern Recognition (CVPR)},
  2020.

\bibitem{Goodfellow.2016}
I.~Goodfellow, Y.~Bengio, and A.~Courville, \emph{Deep learning}, ser. Adaptive
  computation and machine learning.\hskip 1em plus 0.5em minus 0.4em\relax
  Cambridge, Massachusetts and London, England: {The MIT Press}, 2016.

\bibitem{Hassid.07.11.2022}
M.~Hassid, H.~Peng, D.~Rotem, J.~Kasai, I.~Montero, N.~A. Smith, and
  R.~Schwartz, ``How much does attention actually attend?questioning the
  importance of attention in pretrained transformers,'' in \emph{The 2022
  Conference on Empirical Methods in Natural Language Processing (EMNLP)},
  2022.

\bibitem{xu2018how}
{Keyulu Xu and Weihua Hu and Jure Leskovec and Stefanie Jegelka}, ``How
  powerful are graph neural networks?'' in \emph{International Conference on
  Learning Representations}, 2019.

\bibitem{Klimke.642022692022}
M.~Klimke, B.~Volz, and M.~Buchholz, ``Cooperative behavior planning for
  automated driving using graph neural networks,'' in \emph{2022 IEEE
  Intelligent Vehicles Symposium (IV)}, 2022, pp. 167--174.

\bibitem{understandGAT}
{Knyazev, Boris and Taylor, Graham W. and Amer, Mohamed R.}, ``Understanding
  attention and generalization in graph neural networks,'' in \emph{Proceedings
  of the 33rd International Conference on Neural Information Processing Systems
  (NIPS)}, 2019, vol.~33, pp. 4202--4212.

\bibitem{Liu.2022}
Y.~Liu, X.~Qi, E.~A. Sisbot, and K.~Oguchi, ``Multi-agent trajectory prediction
  with graph attention isomorphism neural network,'' in \emph{2022 IEEE
  Intelligent Vehicles Symposium (IV)}, 2022.

\bibitem{Luong.17.08.2015}
\BIBentryALTinterwordspacing
M.-T. Luong, H.~Pham, and C.~D. Manning, ``Effective approaches to
  attention-based neural machine translation.'' [Online]. Available:
  \url{http://arxiv.org/pdf/1508.04025v5}
\BIBentrySTDinterwordspacing

\bibitem{Min.2020}
Y.~Min, F.~Wenkel, and G.~Wolf, ``Scattering gcn: Overcoming oversmoothness in
  graph convolutional networks,'' in \emph{Advances in Neural Information
  Processing Systems}, {H. Larochelle}, {M. Ranzato}, {R. Hadsell}, {M.F.
  Balcan}, and {H. Lin}, Eds., vol.~33, 2020, pp. 14\,498--14\,508.

\bibitem{Mo.2022}
X.~Mo, Z.~Huang, Y.~Xing, and C.~Lv, ``Multi-agent trajectory prediction with
  heterogeneous edge-enhanced graph attention network,'' \emph{IEEE
  Transactions on Intelligent Transportation Systems}, vol.~23, no.~7, pp.
  9554--9567, 2022.

\bibitem{Naik.642022692022}
A.~Y. Naik, A.~Bighashdel, P.~Jancura, and G.~Dubbelman, ``Scene
  spatio-temporal graph convolutional network for pedestrian intention
  estimation,'' in \emph{2022 IEEE Intelligent Vehicles Symposium (IV)}, 2022,
  pp. 874--881.

\bibitem{Neumeier.2021}
M.~Neumeier, M.~Botsch, A.~Tollk{\"u}hn, and T.~Berberich, ``Variational
  autoencoder-based vehicle trajectory prediction with an interpretable latent
  space,'' in \emph{IEEE International Intelligent Transportation Systems
  Conference (ITSC)}, 2021, pp. 820--827.

\bibitem{Neumeier.GFTNN}
M.~Neumeier, A.~Tollk{\"u}hn, M.~Botsch, and W.~Utschick, ``A multidimensional
  graph fourier transformation neural network for vehicle trajectory
  prediction,'' in \emph{2022 IEEE 25th International Conference on Intelligent
  Transportation Systems (ITSC)}, 2022, pp. 687--694.

\bibitem{Park.2017}
D.~H. Park, L.~A. Hendricks, Z.~Akata, A.~Rohrbach, B.~Schiele, T.~Darrell, and
  M.~Rohrbach, ``Attentive explanations: Justifying decisions and pointing to
  the evidence (extended abstract),'' in \emph{arXiv preprint
  arXiv:1711.07373}, 2017.

\bibitem{pytorchGATv2CONV}
\BIBentryALTinterwordspacing
{PyTorch Geometric}, ``Gatv2conv.'' [Online]. Available:
  \url{https://pytorch-geometric.readthedocs.io/en/latest/modules/nn.html#torch_geometric.nn.conv.GATv2Conv}
\BIBentrySTDinterwordspacing

\bibitem{Renz.25.10.2022}
K.~Renz, K.~Chitta, O.-B. Mercea, A.~S. Koepke, Z.~Akata, and A.~Geiger,
  ``Plant: Explainable planning transformers via object-level
  representations,'' in \emph{6th Conference on Robot Learning (CoRL)}, 2022.

\bibitem{Serrano.2019}
S.~Serrano and N.~A. Smith, ``Is attention interpretable?'' in
  \emph{Proceedings of the 57th Annual Meeting of the Association for
  Computational Linguistics}, A.~Korhonen, D.~Traum, and L.~M{\`a}rquez,
  Eds.\hskip 1em plus 0.5em minus 0.4em\relax Stroudsburg, PA, USA:
  {Association for Computational Linguistics}, pp. 2931--2951.

\bibitem{GATv2}
{Shaked Brody, Uri Alon, Eran Yahav}, ``How attentive are graph attention
  networks?'' in \emph{International Conference on Learning Representations
  (ICLR)}, 2022, vol.~10.

\bibitem{Vaswani.2017}
A.~Vaswani, N.~Shazeer, N.~Parmar, J.~Uszkoreit, L.~Jones, A.~N. Gomez,
  L.~Kaiser, and I.~Polosukhin, ``Attention is all you need,'' in
  \emph{Advances in Neural Information Processing Systems}, I.~Guyon, U.~V.
  Luxburg, S.~Bengio, H.~Wallach, R.~Fergus, S.~Vishwanathan, and R.~Garnett,
  Eds., vol.~30, 2017.

\bibitem{Velickovic.30.10.2017}
P.~Veli{\v{c}}kovi{\'c}, G.~Cucurull, A.~Casanova, A.~Romero, P.~Li{\`o}, and
  Y.~Bengio, ``Graph attention networks,'' in \emph{International Conference on
  Learning Representations (ICLR)}, 2018, vol.~6.

\bibitem{Ying.2019}
Z.~Ying, D.~Bourgeois, J.~You, M.~Zitnik, and J.~Leskovec, ``Gnnexplainer:
  Generating explanations for graph neural networks,'' in \emph{Advances in
  Neural Information Processing Systems}, vol.~32, 2019.

\bibitem{Ying.2018}
Z.~Ying, J.~You, C.~Morris, X.~Ren, W.~Hamilton, and J.~Leskovec,
  ``Hierarchical graph representation learning with differentiable pooling,''
  in \emph{Advances in Neural Information Processing Systems}, {S. Bengio}, {H.
  Wallach}, {H. Larochelle}, {K. Grauman}, {N. Cesa-Bianchi}, and {R. Garnett},
  Eds., vol.~31, 2018.

\end{thebibliography}
 }
\end{document}